%% file: main.tex
\documentclass{article}

\usepackage{iclr2027_conference,times}
\usepackage{graphicx}
\usepackage{booktabs}
\usepackage{array}
\usepackage{amsmath,amssymb}
\usepackage{enumitem}
\usepackage{placeins}
\usepackage{float}
\usepackage{microtype}
\usepackage{wrapfig}
\usepackage[hyphens]{url}
\usepackage{xcolor}
\usepackage{colortbl}
\usepackage{tabularx}
\usepackage{caption}
\usepackage[colorlinks=true,
            linkcolor=blue,
            citecolor=blue,
            urlcolor=blue]{hyperref}

\definecolor{riceRow}{HTML}{EEF4FA}
\definecolor{indexRow}{HTML}{F2F2F2}

\input{style/math_commands.tex}

\newcolumntype{P}[1]{>{\raggedright\arraybackslash}p{#1}}
\newcommand{\RICE}{\textsc{Rice-Alpha}}
\newcommand{\MML}{\textsc{Mml}}

\title{RICE-Alpha: \textbf{R}eliability-\textbf{I}nformed\\
\textbf{C}orrection with \textbf{E}vent Graphs for\\
LLM-Agent Stock Forecasting}
\hypersetup{pdftitle={RICE-Alpha: Reliability-Informed Correction with Event Graphs for LLM-Agent Stock Forecasting}}

\input{authors.tex}
\author{\RICEAuthors}
\iclrfinalcopy

\begin{document}

\maketitle
\lhead{Preprint}

\begin{abstract}
Equity-relevant news evolves through temporally dependent corporate events rather than isolated observations, making historical information useful only when event continuity, information availability, and transition reliability are modeled explicitly. Existing LLM-based financial agents increasingly incorporate historical evidence, yet they provide limited support for preserving issuer-specific chronology under point-in-time constraints and for identifying when historical transitions contribute information beyond the current forecast. We present \RICE{} (Reliability-Informed Correction with Event Graphs), a point-in-time stock-scoring framework that separates a history-aware multi-view Base Alpha from a reliability-calibrated residual correction derived from historical event continuation. A Multi-Tier Memory Layer grounds news interpretation in temporally eligible issuer-specific history, while a Typed Event Agent constructs event states whose successor relations are formed within issuers and pooled across firms only after valid local pairing. Matured transitions are calibrated by their empirical reliability, and the resulting graph signal is residualized against the Base Alpha and technical view to obtain the RICE Delta. On daily Nasdaq-100 and Hang Seng Index panels from 2024 to 2026, \RICE{} achieves the strongest results among the evaluated LLM-based agents and momentum across four predictive and four portfolio-level metrics in both markets. Its ICIR more than doubles that of the strongest baseline, while net Sharpe ratios reach \(1.656\) and \(1.725\) in the U.S. and Hong Kong, respectively. U.S. ablations further show significant reductions in IC and RankIC after Holm adjustment when major components are removed. These results indicate that historical event continuation adds incremental information when it is temporally grounded, reliability-calibrated, and introduced as a residual correction to a multi-view forecast.

\end{abstract}

\input{sections/01_intro}
\input{sections/02_related_work}
\input{sections/03_method}
\input{sections/04_experiments}
\input{sections/05_conclusion}
\input{sections/06_limitations}

\begingroup
\raggedright
\setlength{\bibsep}{0pt}
\bibliography{references}
\bibliographystyle{style/iclr2027_conference}
\endgroup

\clearpage
\appendix
\input{appendix/definitions}

\end{document}

%% file: style/math_commands.tex
\usepackage{amsmath,amsfonts,bm}

\def\eqref#1{equation~\ref{#1}}

\def\1{\bm{1}}

\DeclareMathAlphabet{\mathsfit}{\encodingdefault}{\sfdefault}{m}{sl}
\SetMathAlphabet{\mathsfit}{bold}{\encodingdefault}{\sfdefault}{bx}{n}



%% file: authors.tex
\hypersetup{pdfauthor={Tong Liu, Lanmiao Liu, Xiang Hu}}
\newcommand{\RICEAuthors}{%
  Tong Liu$^{1,*}$\And
  Lanmiao Liu$^{2,3,*}$\And
  Xiang Hu$^{4,*,\dagger}$\AND
  \normalfont\small
  $^1$Zircon Security \qquad $^2$Utrecht University\\
  $^3$The Max Planck Institute for Psycholinguistics\\
  $^4$China Life R\&D Center\\
  $^*$Equal contribution. \quad
  $^\dagger$Corresponding author: \texttt{huxiang2022@e-chinalife.com}\\
  Tong Liu: \texttt{liutong@zr.hk}\\
  Lanmiao Liu: \texttt{lanmiao.liu@mpi.nl}%
}

%% file: sections/01_intro.tex
\section{Introduction}
\label{sec:intro}

Financial decision-making requires integrating heterogeneous information from firm-level news \citep{boudoukh2019information}, financial statements \citep{ou1989financial}, market dynamics \citep{garleanu2013dynamic}, and macroeconomic conditions \citep{fama1989business}. Concurrent advances in large language model (LLM) architectures \citep{liu2024deepseek,achiam2023gpt}, training paradigms, and data scaling have substantially improved their ability to represent and integrate heterogeneous information, providing a natural foundation for financial reasoning. Recent studies leverage LLMs to extract predictive signals from time-sensitive financial text and combine them with structured market information \citep{xiong2025flag,lopez2026can}. Financial information, however, evolves continuously: newly observed news may update an existing corporate event, interact with prior issuer-specific developments, or acquire a different interpretation under changing market conditions. Effective financial reasoning therefore requires not only cutoff-available evidence, but also a temporally valid representation of how that evidence relates to historical event dynamics.

Related design challenges appear in other generative-AI domains: a survey of fashion generation organizes heterogeneous image, 3D, and video tasks \citep{shi2025generative}; FonTS and WordCon study fine-grained control of rendered text \citep{shi2025fonts,11494032}; and vision-guided audio alignment supports multimodal emotion understanding \citep{zhang2025learninghearseeingits}. AnySurf uses directed edges to represent surface orientation in 3D generation \citep{shi2026anysurfsurfacegenerationdirected}. These works provide cross-domain examples of structured representations and controlled information integration, motivating our attention to explicit event states and directed transitions in financial forecasting.

LLM-based financial agents increasingly retrieve historical information to contextualize current evidence \citep{yu2024finmem,guo2026meme,qian2026sleipnir}, while event-based forecasting methods exploit structured news relations and temporal dependencies for return prediction \citep{han2025structured,li2024causalstock,li2026multi}. Historical context and event structure, however, remain difficult to use reliably under a strict point-in-time setting. Retrieved evidence must remain consistent with what was observable at the prediction cutoff, while event relations should preserve issuer-specific chronology rather than arise from spurious cross-stock pairing. Structured event models show lag-dependent, time-varying inter-stock relations \citep{li2024causalstock,liu2024echo}, suggesting that the predictive relevance of historical transitions varies with empirical support, temporal stability, market context, and association with subsequent returns. These observations expose three coupled challenges in history-aware financial reasoning: \textbf{limited historical contextualization}, \textbf{weak temporal grounding of event relations}, and \textbf{uncalibrated use of historical transitions}.

\begin{wrapfigure}{r}{0.5\linewidth}
    \centering
    \vspace{-0.5em}
    \includegraphics[width=\linewidth]{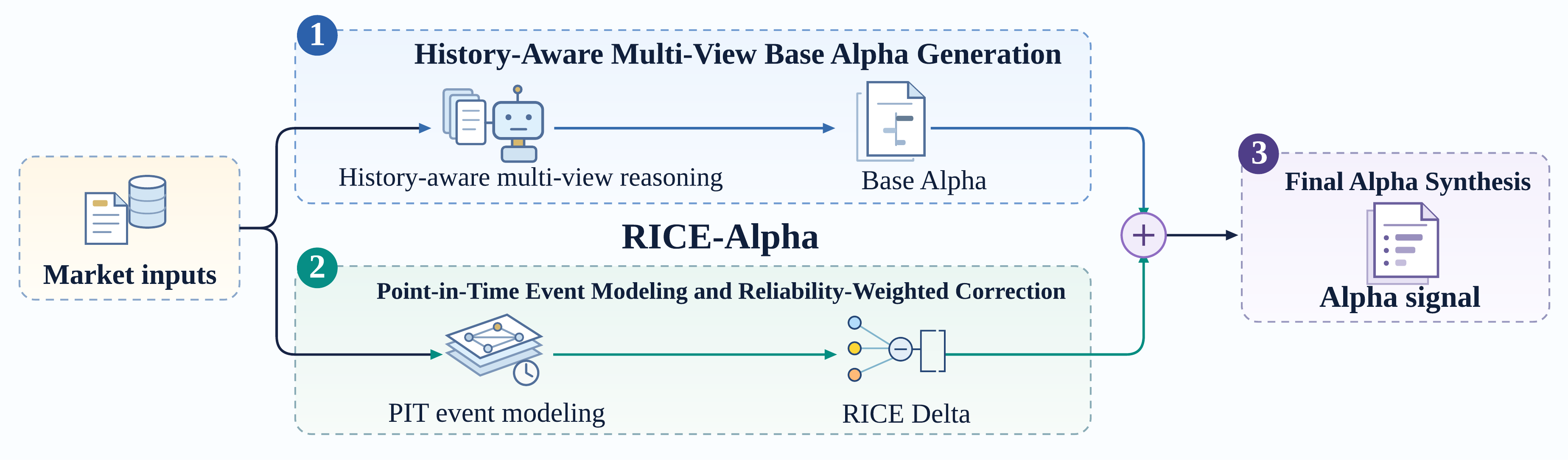}

    \captionsetup{font=small}
    \caption{Overview of the RICE-Alpha framework. A point-in-time multi-view Base Alpha is refined by a reliability-calibrated residual derived from historical event propagation.}
    \label{fig:rice_intro}

    \vspace{-0.8em}
\end{wrapfigure}

To address these challenges, we present \RICE{} (\emph{Reliability-Informed Correction with Event Graphs}), a point-in-time stock-scoring framework that separates a history-aware multi-view reference signal from the incremental contribution of historical event continuation. As illustrated in Figure~\ref{fig:rice_intro}, \RICE{} follows three stages. First, \emph{History-Aware Multi-View Base Alpha Generation} constructs a point-in-time reference signal through dedicated sentiment, technical, fundamental, and macroeconomic agents. A structured Multi-Tier Memory Layer (\MML{}) grounds the Sentiment Agent in temporally eligible issuer-specific history, while the remaining agents capture complementary cutoff-available market views. The resulting stock-level signals are reconciled before bounded macro adjustment to obtain Base Alpha. Second, \emph{Point-in-Time Event Modeling and Reliability-Weighted Correction} is performed by the RICE Event Engine. A Typed Event Agent converts cutoff-available news and prior active records into structured event states. Historical successor relations are formed strictly within each issuer and pooled across firms only after valid local pairing, yielding a frozen point-in-time typed-event graph. Current states retrieve eligible one-hop transitions whose contributions are calibrated by transition frequency, regime-relative lift, lag structure, matured successor-return association, and observational reliability. The resulting graph signal is residualized against the Base Alpha and technical view to form the RICE Delta. Finally, \emph{Final Alpha Synthesis} combines Base Alpha and RICE Delta through a bounded residual composition.

We evaluate \RICE{} on daily U.S. and Hong Kong equity panels using four predictive metrics and four portfolio-level backtesting metrics. The empirical comparison focuses on LLM-agent stock-scoring systems and a price-based momentum baseline under a common evaluation protocol, while prior event-graph methods serve as methodological antecedents rather than directly benchmarked systems. Across both markets, \RICE{} achieves the strongest results among the evaluated methods on all eight metrics. Ablation studies on the U.S. panel show significant reductions in IC and RankIC after Holm correction when major components are removed. Evaluations with an alternative LLM backbone, constituent-exclusion analysis, transaction-cost stress tests, and regime-specific analysis further examine robustness across modeling and trading conditions.

Our contributions are summarized as follows:
\begin{itemize}
    \item \textbf{History-Aware Multi-View Base Alpha Generation.} We introduce a point-in-time multi-agent formulation that integrates sentiment, technical, fundamental, and macroeconomic evidence while grounding the Sentiment Agent in temporally eligible issuer-specific history through a structured \MML{}. This enables historical context to inform current-news reasoning under a consistent point-in-time information set.

    \item \textbf{Point-in-Time Issuer-Local Event Graph Construction.} We introduce a Typed Event Agent that maps cutoff-available news to structured event records and typed states, and construct successor relations strictly within each issuer. Comparable transitions are pooled across issuers only after valid local pairing, preserving issuer-specific chronology while preventing look-ahead and spurious cross-stock transitions.

    \item \textbf{Reliability-Calibrated Residual Event Correction.} We develop a reliability-aware mechanism that converts matured one-hop transitions into a separate RICE Delta using transition frequency, regime-relative lift, lag structure, successor-return association, and observational reliability. The graph signal is residualized against the Base Alpha and technical view before final synthesis, isolating the incremental predictive content of historical event continuation.
\end{itemize}

%% file: sections/02_related_work.tex
\section{Related Work}
\label{sec:related_work}

\paragraph{LLM-Based Agents for Financial Trading.}
Financial prediction has been studied through reinforcement learning and market representation \citep{wang2021deeptrader,liu2022finrlmeta,li2024master}, formulaic alpha discovery \citep{yu2023alphagen,shi2025alphaforge,zhu2025alphaqcm,chen2026alphasage,shi2026alphajungle}, and financial language models for textual reasoning and return forecasting \citep{araci2019finbert,wu2023bloomberggpt,yang2023fingpt,xiao2024tradingagents,xie2023wallstreetneophyte}. Recent agentic systems further automate factor and strategy development \citep{zhang2026alphaforgebench,li2025rdagentquant,song2026timi}, while layered-memory and experience-retrieval mechanisms incorporate historical context into LLM-based decision making \citep{yu2024finmem,zhao2024expel,fountas2025emllm}. Despite these advances, current market views and retrieved historical evidence are often processed through separate reasoning or signal-generation pathways \citep{yu2024finmem,xiao2024tradingagents,song2026timi}, providing limited support for grounding current news in issuer-specific event history before cross-view aggregation. \RICE{} differs by conditioning the sentiment view on temporally eligible historical evidence through a structured \MML{} within a point-in-time multi-view Base Alpha.

\paragraph{Temporal Event Modeling and Reliability-Aware Forecasting.}
Structured event representations encode temporal, semantic, and causal dependencies in financial news \citep{han2025structured,li2024causalstock}, while graph-based methods extend this formulation to cross-stock interactions, supply-chain relations, and dynamically evolving market structures \citep{qian2024mdgnn,huang2026crossstock,yilki2026supplychain,zhang2026dynamicfinancialkg,liu2024echo}. TRACE \citep{ding2026trace} combines temporal graph structure with mined rules and multi-hop reasoning, while retrieval- and memory-based systems incorporate historical availability, delayed feedback, and evolving contextual state \citep{zhao2026pointintime,deng2026treeexperience,yu2024finmem,fountas2025emllm}. These studies establish temporal event structure as an important source of predictive context. Relation construction over pooled observations, however, can attenuate issuer-specific chronology \citep{li2024causalstock,huang2026crossstock}, whereas point-in-time forecasting requires queried relations, outcomes, and memory states to remain restricted to information available at the prediction cutoff \citep{zhao2026pointintime,deng2026treeexperience}. \RICE{} differs in its construction principle by forming successor relations within each issuer and pooling comparable transitions only after valid local pairing under point-in-time constraints.

Graph-based stock models learn time-varying, lag-dependent relation strengths \citep{qian2024mdgnn,liu2024echo,li2024causalstock}, so historical transitions may differ in predictive relevance with empirical support, market regime, and return stability. Financial-agent frameworks increasingly decouple language-based reasoning from downstream numerical execution \citep{zhang2026alphaforgebench,li2025rdagentquant,song2026timi,koa2026vta}, while structured consensus and temporal evaluation improve the stability of intermediate reasoning \citep{guo2026meme}. These event-graph and event-reasoning approaches are methodologically related to \RICE{} but are not directly benchmarked under the empirical protocol used in this study. Accordingly, our distinction is architectural rather than an empirical claim of superiority over prior event-graph methods. \RICE{} represents matured transition evidence through a separate reliability-weighted residual, allowing its incremental contribution relative to the point-in-time Base Alpha to be isolated before final synthesis.

%% file: sections/03_method.tex
\section{Methodology}
\label{sec:method}

\subsection{Overview of RICE-Alpha}
\label{sec:method_overview}

\begin{figure}[tb]
  \centering
  \vspace{-0.5em}
  \begingroup
  \setlength{\linewidth}{0.85\linewidth}
  \pdfximage width\linewidth cropbox{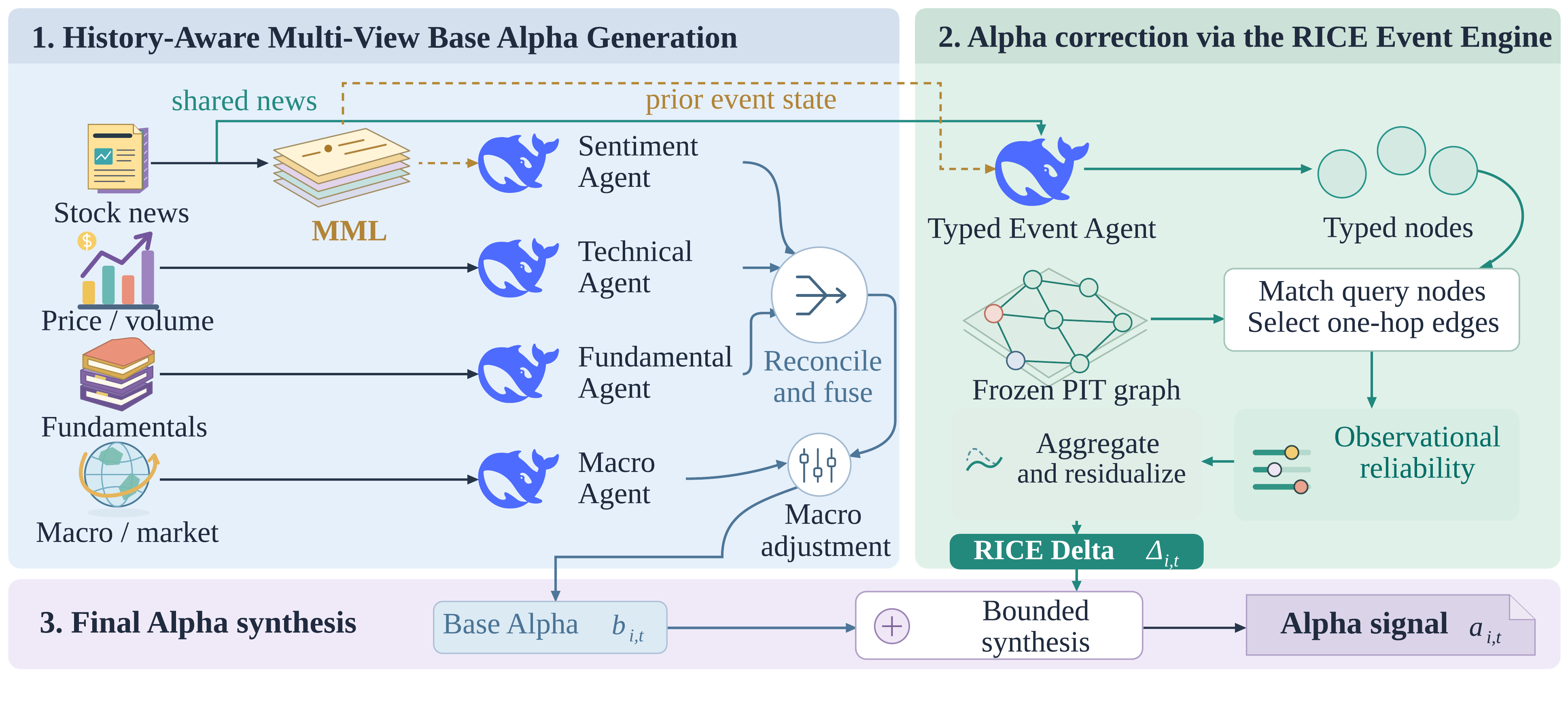}
  \leavevmode\pdfrefximage\pdflastximage
  \endgroup
  \caption{Overview of the \RICE{} pipeline. Base Alpha captures multi-view evidence, RICE Delta captures reliability-weighted event continuation, and their bounded synthesis yields the final stock score.}
  \label{fig:rice_overview}
\end{figure}

As illustrated in Figure~\ref{fig:rice_overview}, we present \RICE{} (\emph{Reliability-Informed Correction with Event Graphs}), a point-in-time stock-scoring framework that separates a history-aware multi-view reference signal from an explicit correction derived from historical event continuation. \RICE{} comprises three modules: \emph{History-Aware Multi-View Base Alpha Generation}, where dedicated sentiment, technical, fundamental, and macroeconomic agents process complementary financial views, with \MML{} conditioning the Sentiment Agent on temporally eligible issuer-specific history before reconciliation into Base Alpha; \emph{Alpha Correction via the RICE Event Engine}, which maps current and prior event records to typed states, queries a frozen PIT graph, and aggregates reliability-weighted one-hop transition evidence into RICE Delta; and \emph{Final Alpha Synthesis}, which combines the two signals through a bounded residual composition. Cutoff-available news is therefore used along two complementary paths: \MML{} supports history-aware semantic interpretation, while the Typed Event Agent and frozen event graph provide structured numerical evidence for event-continuation correction.

At signal date \(t\), given tradable universe \(\mathcal U_t\) and cutoff-available news, price--volume history, fundamentals, and macro/market indicators, \RICE{} outputs a cross-sectional stock-ranking score \(a_{i,t}\) for each \(i\in\mathcal U_t\). Evaluation follows a \(T{+}1\) execution convention.

\vspace{-0.8em}
\subsection{History-Aware Multi-View Base Alpha Generation}
\label{sec:base_alpha}

For stock \(i\) at signal date \(t\), the input state is \(\mathcal X_{i,t}=(\mathcal N_{i,t},\mathcal T_{i,t},\mathcal F_{i,t},\mathcal M_t)\), where \(\mathcal N_{i,t}\) denotes cutoff-available stock news, \(\mathcal T_{i,t}\) the price--volume history observed through \(t\), \(\mathcal F_{i,t}\) the admitted fundamental information, and \(\mathcal M_t\) the dated macro/market indicators. The Sentiment, Technical, and Fundamental Agents process \(\mathcal N_{i,t}\), \(\mathcal T_{i,t}\), and \(\mathcal F_{i,t}\), respectively, and return view-specific signals with associated confidence scores, while the Macro Agent maps \(\mathcal M_t\) to a shared date-level market signal.

The Sentiment Agent is additionally conditioned on the Multi-Tier Memory Layer (\MML{}), which retrieves temporally eligible issuer-specific context, including local graph context, matured historical cases, the current event narrative, regime-matched calibration evidence, and direct successor-event memories. Outcome-dependent memory is admitted only after the corresponding returns have matured. \MML{} supports semantic interpretation of current news, whereas numerical transition statistics are handled separately by the RICE Event Engine in Section~\ref{sec:rice_engine}.

Let \(\mathbf v_{i,t}=(v_{i,t}^{S},v_{i,t}^{T},v_{i,t}^{F})\) denote the standardized sentiment, technical, and fundamental signals, and let \(\widetilde{\mathbf w}_{i,t}\) denote their effective weights after confidence filtering, mature-feedback adjustment, and disagreement handling. Their weighted reconciliation defines \(\widetilde v_{i,t}\), while the Macro Agent induces a stock-specific adjustment \(m_{i,t}\). The Base Alpha is
\begin{equation}
b_{i,t}
=
\operatorname{clip}\!\left(
m_{i,t}\widetilde v_{i,t},
b_{\min},
b_{\max}
\right),
\label{eq:base_alpha}
\end{equation}
where \(b_{\min}\) and \(b_{\max}\) define the admissible Base Alpha range. Both \(\widetilde v_{i,t}\) and \(m_{i,t}\) are constructed exclusively from information available before the prediction cutoff. The resulting \(b_{i,t}\) serves as the reference signal for the residual correction in Section~\ref{sec:rice_engine}.

\FloatBarrier

\subsection{Alpha Correction via the RICE Event Engine}
\label{sec:rice_engine}

\begin{figure}[!htbp]
  \centering
  \vspace{-0.8em}
  \begingroup
  \setlength{\linewidth}{0.85\linewidth}
  \pdfximage width\linewidth cropbox{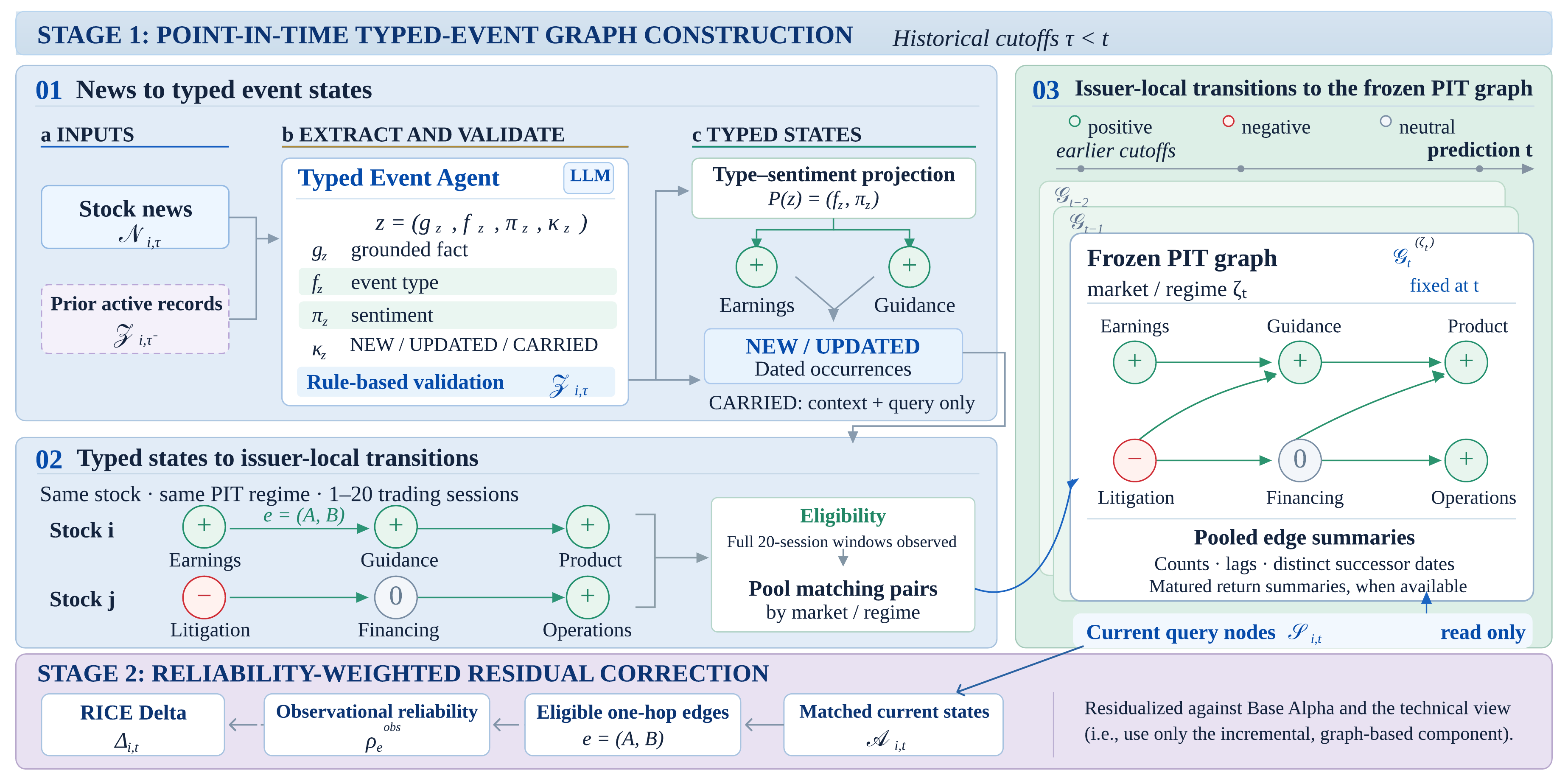}
  \leavevmode\pdfrefximage\pdflastximage
  \endgroup
  \caption{Overview of the RICE Event Engine. Stage~1 constructs a frozen point-in-time typed-event graph from issuer-local successor relations, while Stage~2 converts reliability-weighted graph evidence into the residual RICE Delta \(\Delta_{i,t}\).}
  \label{fig:rice_typed_event_graph}
  \vspace{-0.8em}
\end{figure}

Conditioned on the Base Alpha \(b_{i,t}\), the RICE Event Engine models predictive structure arising from the temporal evolution of corporate events. Issuer-specific news events form recurring successor patterns whose statistical regularities may persist across firms and market regimes. Semantic event extraction is performed by the Typed Event Agent, while graph construction, reliability weighting, aggregation, and residualization are deterministic. Figure~\ref{fig:rice_typed_event_graph} details its two-stage construction.

\subsubsection{Stage 1: Point-in-Time Typed-Event Graph Construction}

\paragraph{News to typed event states.}
Corporate news streams contain both genuine event progression and repeated reporting, so treating individual articles as independent events would distort transition statistics. At each historical cutoff \(\tau<t\), the Typed Event Agent jointly processes stock news \(\mathcal N_{i,\tau}\) and the previously validated active record set \(\mathcal Z_{i,\tau^-}\), resolving newly reported information against existing issuer-specific episodes. Each accepted record is \(z=(g_z,f_z,\pi_z,\kappa_z)\), where \(g_z\) is a grounded event description, \(f_z\) an event type, \(\pi_z\) an event-level sentiment label, and \(\kappa_z\in\{\textsc{New},\textsc{Updated},\textsc{Carried}\}\) the lifecycle state. \textsc{New} opens an unmatched episode, \textsc{Updated} records substantive information about an active episode, and \textsc{Carried} preserves an unresolved episode without introducing a new occurrence. Only validated \textsc{New} and \textsc{Updated} records create dated graph occurrences, preventing repeated coverage from inflating transition support. Each occurrence is projected to \(P(z)=(f_z,\pi_z)\), while episode identity and lifecycle metadata remain attached to the underlying record. At prediction cutoff \(t\), typed states associated with current validated records, including \textsc{Carried} episodes, form the query set \(\mathcal S_{i,t}\).

\paragraph{Typed states to issuer-local transitions.}
Temporal succession is estimated within issuers to distinguish event evolution from incidental co-occurrence. For two eligible occurrences of the same stock with typed states \(A\) and \(B\), Stage~1 admits \(A\!\rightarrow\!B\) when \(B\) occurs within the successor window \(W_{\mathrm{succ}}\) after \(A\) and both occurrences belong to the same market regime \(\zeta\). Issuer-local pairing preserves corporate-event chronology, while regime consistency separates successor patterns observed under distinct market conditions. Each directed pair represents temporal order rather than causation. An occurrence becomes an eligible anchor only after its full successor window has elapsed, preventing incomplete follow-up from being interpreted as absence of successors.

\paragraph{Issuer-local transitions to the frozen PIT graph.}
Individual firms typically provide too few repeated transitions for stable estimation, so Stage~1 pools comparable directed pairs across firms only after issuer-local formation and within the same market and regime \(\zeta\). The graph at cutoff \(t\) is \(\mathcal G_t^{(\zeta)}=(\mathcal V_t^{(\zeta)},\mathcal E_t^{(\zeta)})\), where \(A=(f,\pi)\in\mathcal V_t^{(\zeta)}\) denotes a typed event state and \(e=(A,B)\in\mathcal E_t^{(\zeta)}\) a matured successor relation. This local-first, pool-second construction combines issuer-specific chronology with cross-sectional statistical strength. Each pooled edge summarizes transition support, regime-relative frequency, lag structure, distinct successor dates, and matured successor-return behavior. The graph is frozen at cutoff \(t\), so current records may query but cannot update the statistics used for their own prediction.

\subsubsection{Stage 2: Reliability-Weighted Residual Correction}

\paragraph{Current states to one-hop successor edges.}
At cutoff \(t\), the current query set is matched against the frozen graph for regime \(\zeta_t\), yielding \(\mathcal A_{i,t}=\mathcal S_{i,t}\cap\mathcal V_t^{(\zeta_t)}\). For each matched state \(A\), Stage~2 considers eligible one-hop outgoing edges \(e=(A,B)\), restricting propagation to directly observed successor structure. Let \(n_A\) denote the number of eligible anchors of \(A\), \(n_e\) the number followed by \(B\), \(n_{\rightarrow B}\) the number of regime-matched anchors followed by \(B\), and \(N_{\zeta}\) the total number of eligible anchors in regime \(\zeta\). The smoothed conditional transition rate \(p_e=(n_e+\alpha_p)/(n_A+\alpha_p+\beta_p)\) estimates how frequently \(B\) follows \(A\), while \(q_e=(n_{\rightarrow B}+\alpha_q)/(N_{\zeta}+\alpha_q+\beta_q)\) provides the corresponding regime baseline. Their ratio \(L_e=p_e/\max(q_e,\varepsilon)\) defines relative lift, discounting broadly prevalent successors. Candidate edges are restricted to non-self transitions and filtered by support, transition strength, and per-node budget constraints.

\paragraph{Successor edges to reliability weights.}
Transition recurrence alone is insufficient to establish predictive relevance. Each retained edge therefore receives an observational reliability weight \(\rho_e^{\mathrm{obs}}\in[0,1]\) based on matured transition support, coverage across distinct successor dates, and standardized successor-return association; lag alignment enters the edge contribution separately. Successor returns are admitted only after maturity, adjusted within market and date, and aggregated at the successor-date level to prevent repeated observations from dominating the estimate.

\paragraph{Weighted edges to the residual RICE Delta.}
Each retained edge \(e=(A,B)\) contributes a bounded score \(C_{i,e,t}\) whose sign follows the matured successor-return association and whose magnitude combines transition strength, observational reliability, query-state importance, and temporal alignment. Aggregating retained outgoing edges yields
{\small
\begin{equation}
r_{i,t}^{\mathrm{graph}}
=
\operatorname{clip}\!\left(
\sum_{A\in\mathcal A_{i,t}}
\sum_{e\in\mathcal E_t^{(\zeta_t),\rightarrow}(A)}
C_{i,e,t},
-\tau_{\Delta},
\tau_{\Delta}
\right),
\label{eq:raw_graph_signal}
\end{equation}
}
where \(\mathcal E_t^{(\zeta_t),\rightarrow}(A)\) denotes the retained outgoing edges of \(A\) and \(\tau_{\Delta}\) bounds the aggregate graph contribution. Because \MML{} and the technical view may already encode information associated with the same underlying events, direct addition can duplicate information contained in the reference forecast. The graph signal is therefore residualized against the Base Alpha and technical view:
\begin{equation}
\Delta_{i,t}
=
g_{i,t}^{\mathrm{edge}}
\,
\mathsf{Res}_{t}\!\left(
r_{\cdot,t}^{\mathrm{graph}};
b_{\cdot,t},
v_{\cdot,t}^{T}
\right)_i,
\label{eq:rice_delta}
\end{equation}
where \(g_{i,t}^{\mathrm{edge}}\in\{0,1\}\) indicates whether stock \(i\) has at least one retained non-self transition. The gating restricts correction to observations supported by matched historical evidence, while \(\mathsf{Res}_{t}\) removes cross-sectional rank-linear overlap with the controls before standardization.

\subsection{Final Alpha Synthesis}
\label{sec:prediction}

The RICE Delta \(\Delta_{i,t}\) is combined with Base Alpha \(b_{i,t}\) through a bounded residual composition:
\begin{equation}
a_{i,t}
=
\operatorname{clip}\!\left(
b_{i,t}
+
\lambda_{\Delta}s_{\Delta}\Delta_{i,t},
a_{\min},
a_{\max}
\right),
\label{eq:final_prediction}
\end{equation}
where \(s_{\Delta}>0\) aligns the standardized residual correction with the Base Alpha scale, \(\lambda_{\Delta}\) controls its contribution, and \(a_{\min}\) and \(a_{\max}\) define the admissible score range. The indispensable parameter and residualization definitions are collected in Appendix~\ref{app:definitions}.

%% file: sections/04_experiments.tex
\section{Experiments and Results}
\label{sec:experiments}

\subsection{Experimental Settings}
\label{sec:evaluation_contract}

\paragraph{Task, data, and baselines.}
On each signal date, every method scores the Nasdaq-100 (U.S.) or Hang Seng Index (Hong Kong) constituents against the five-session return from the next open, using date-specific membership, adjusted prices, financials, news (English and Chinese in Hong Kong), and market indicators \citep{alphavantageDocumentation,tushareDocumentation}. Following a 2023 warm-up that initializes memory and event statistics, evaluation runs from 2024-01-02 to 2026-03-30 (562 U.S. and 551 Hong Kong signal dates). LLM-agent baselines, MEME \citep{guo2026meme}, R\&D-Agent-Quant \citep{li2025rdagentquant}, and AI Hedge Fund \citep{aihedgefund}, test whether \RICE{} adds value beyond existing agentic systems; price-only 12--1 momentum tests it against a simple trend signal. All methods share each market's data, period, targets, and dated universe. \RICE{} uses DeepSeek-V4-Flash \citep{deepseekai2026deepseekv4} without fine-tuning.

\paragraph{Portfolio and frozen configuration.}
Every strategy in Table~\ref{tab:main_results} follows one long-only rule, so strategies differ only in their scores. At each weekly rebalance, the portfolio is reset to equal weights on the top-scoring 10\% of constituents, trades at the next open, and is held until the next rebalance. It pays trading costs of 4+4~bp per side in the U.S. and 10+4~bp in Hong Kong (commission plus slippage). Prompts, hyperparameter values, update rules, the portfolio rule, and trading costs were frozen at the end of the 2023 warm-up, with identical hyperparameters in the two markets. No setting was tuned, selected, or revised using 2024--2026 data. During evaluation, memory, event statistics, feedback, and view weights change only through the frozen update rules.

\paragraph{Metrics and tests.}
IC and RankIC (daily cross-sectional correlations of scores with realized returns) measure ranking quality, ICIR and RankICIR day-to-day stability, and ARR, Sharpe, maximum drawdown (MDD), and Calmar ratio (CR) net tradable outcomes. Newey--West \(t\)-statistics (five lags) handle the overlapping five-session targets, and one-sided paired 20-session block bootstraps (5{,}000 replications) with Holm adjustment compare the complete system with the ablation variants and AI Hedge Fund.

\subsection{Main Results}
\label{sec:results}
\input{tables/table1_main_ic}

\paragraph{\RICE{} ranks first on every metric in both markets.}
As shown in Table~\ref{tab:main_results}, \RICE{} supplies the strongest ranking signal in both markets, with the most notable improvement being its repeatability. Compared to momentum (the strongest baseline), IC rises from 0.0316 to 0.0350 in the U.S. and from 0.0160 to 0.0283 in Hong Kong, whereas the best agent baselines reach at most 0.0216 and 0.0013. ICIR more than doubles in both markets (0.2928 vs.\ 0.1274; 0.1986 vs.\ 0.0645), largely because the daily U.S. IC of \RICE{} exhibits half the dispersion (SD 0.120 vs.\ 0.248), consistent with the event correction turning sparse historical episodes into a steadier daily ranking.

\paragraph{The advantage carries over to Hong Kong.}
Hong Kong is a demanding stress test: with the same frozen hyperparameters, it changes the language mix, disclosure sources, and market microstructure. MEME, R\&D-Agent-Quant, and AI Hedge Fund degrade severely, reaching Hong Kong ICs of 0.0013, 0.0011, and \(-0.0018\) (at most 6\% of their U.S. values), while momentum drops from 0.0316 to 0.0160. In contrast, \RICE{} retains 81\% of its U.S. IC, and its Hong Kong RankIC (0.0395) exceeds its U.S. value (0.0360), the only LLM-based method whose signal stays largely intact across markets.

\paragraph{Statistical significance.}
Mean IC and RankIC are significantly positive in both markets (Newey--West \(t\) of 4.31 and 4.24 in the U.S., 2.48 and 3.10 in Hong Kong), whereas those of momentum and AI Hedge Fund are not significant in either market, and paired tests against AI Hedge Fund yield Holm-adjusted \(p \le 0.0096\) (Table~\ref{tab:significance}).

\begin{figure}[!hb]
  \centering
  \includegraphics[width=0.82\linewidth]{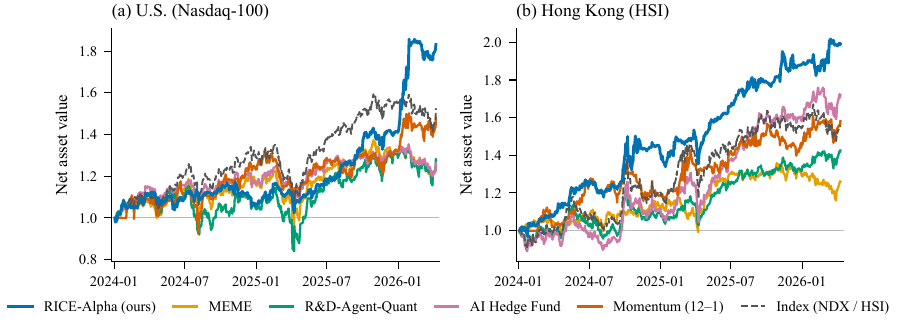}
  \caption{Net asset value of \RICE{}, the baselines, and the index in (a) the U.S.
  and (b) Hong Kong from 2024-01-03; models are net of costs, indices exclude costs.}
  \label{fig:nav}
\end{figure}

\paragraph{Better rankings translate into superior portfolios.}
\RICE{} achieves the highest ARR, Sharpe, and CR and the smallest MDD in each market, with Sharpe ratios of 1.656 in the U.S. and 1.725 in Hong Kong (best baselines: 0.889 and 1.130). Its net ARR exceeds the indices by 10.20 and 14.40 points with roughly half the drawdown (12.09\% vs.\ 24.37\%; 11.83\% vs.\ 21.24\%). Figure~\ref{fig:nav} shows how this edge accumulates: the Hong Kong strategy leads all baselines from late September 2024, while the U.S. strategy limits losses in the 2025 sell-off and pulls ahead in early 2026.

\subsection{Ablation Study}
\label{sec:economic_results}

Table~\ref{tab:ablation_results} introduces the components cumulatively on the U.S. panel, mapping each to a challenge of Section~\ref{sec:intro}: retrieved memory (C1) addresses limited historical contextualization; numerical evidence from the issuer-local event graph (C2) resolves weak temporal grounding; and reliability weighting with residualization (C3) mitigates uncalibrated transitions. The \emph{w/o C3} row injects the graph evidence as a direct correction without either C3 operation, two rows retain exactly one operation each to disentangle reliability weighting from residualization, and a separate run replaces the specialized sentiment, technical, and fundamental agents with one generalist stock-analysis agent.

\input{tables/table2_ablation}
\input{tables/table3_significance}

\paragraph{Every component adds measurable signal.}
Both ranking quality and stability improve monotonically as components are added: IC rises from 0.0064 (no components) to 0.0191 (memory), 0.0260 (direct correction), and 0.0350 (full correction), and ICIR climbs from 0.0472 to 0.1469, 0.2203, and 0.2928, with RankIC and RankICIR following the same trajectory. All five ablation variants trail the complete system on IC and RankIC under six-row Holm tests (maximum adjusted \(p=0.0480\)). Even with memory alone, the ICIR (0.1469) exceeds every baseline.

\paragraph{Reliability and residualization play complementary roles.}
Reliability weighting alone achieves an IC of 0.0326 (ICIR 0.2706), already surpassing momentum, and exceeds residualization alone (0.0208) with unadjusted one-sided \(p=0.0050\) for IC and 0.0406 for RankIC (Table~\ref{tab:significance}, Panel C). Adding residualization on top of reliability weighting raises IC further to 0.0350 and ICIR to 0.2928, a Holm-significant gain. Reliability weighting thus acts as the primary safeguard against noisy continuations, and residualization adds value on top of it by excising overlap with the Base Alpha.

\paragraph{Multi-agent versus single-agent scoring.}
The single-agent run reaches IC 0.0148 and RankIC 0.0142. On the common U.S. panel, the complete system exceeds it by 0.0202 in IC and 0.0218 in RankIC (two-sided block bootstrap, \(p=0.020\) and 0.042), so the comparison favors the complete multi-agent configuration over the evaluated single-agent implementation.

\paragraph{Robustness across backbone, size, and cost.}
With Qwen3-Max in every LLM role, U.S. IC is 0.0281 and ICIR 0.1958, above every agent baseline (at most 0.0216 and 0.1135). Dropping ten large caps keeps IC and RankIC significant in both markets (Newey--West \(t\ge2.69\)), and per-side costs of 20 and 28~bp leave Sharpe ratios of 1.551 and 1.659, above every default-cost baseline.

\subsection{Behavior in Falling Markets}
\label{sec:market_conditions}

Table~\ref{tab:market_conditions} compares all methods on index-down and index-up sessions. Because all strategies share one portfolio rule, gaps between them reflect stock selection, whereas the gap to the index partly comes from the rule itself.

\input{tables/table4_market_conditions}

\paragraph{\RICE{} cuts losses more than gains.}
On index-down sessions, \RICE{} captures 51\% (U.S.) and 34\% (Hong Kong) of the index decline, versus 83\% and 58\% for momentum and 73\% and 93\% for AI Hedge Fund, while retaining 65\% and 51\% of up-session gains. MEME and R\&D-Agent-Quant lose less on ordinary down sessions owing to low market exposure (U.S. beta 0.20 and 0.18), but capture only a third of rallies and still lost 20.8\% and 24.9\% in the Nasdaq-100's largest crash (index 24.4\%), whereas \RICE{} lost 10.9\%, the least of all methods. In months when the index fell, \RICE{} posted the best mean return in both markets (\(-0.4\%\) and \(+1.3\%\)) and outperformed the index in 6 of 7 U.S. months and all 13 Hong Kong months.

%% file: tables/table1_main_ic.tex
\begin{table}[!hb]
\centering
\caption{Main results on common data and dated universes. ICIR and RankICIR are not annualized; ARR and MDD (a positive loss) are in \%. Bold: market best. All strategies use the long-only, equal-weighted top-10\% weekly rule and costs of Section~\ref{sec:evaluation_contract}; NDX/HSI exclude costs. $^{\dagger}$/$^{\ddagger}$: lag-5 Newey--West $t>1.96$/$2.58$.}
\label{tab:main_results}
\scriptsize
\setlength{\tabcolsep}{2pt}
\renewcommand{\arraystretch}{0.90}
\begin{tabularx}{\linewidth}{@{}>{\raggedright\arraybackslash}Xcccccccc@{}}
\toprule
& \multicolumn{4}{c}{U.S. (Nasdaq-100)} & \multicolumn{4}{c}{Hong Kong (HSI)} \\
\cmidrule(lr){2-5}\cmidrule(l){6-9}
\multicolumn{9}{l}{\textit{Panel A: Prediction quality}} \\
Method & IC $\uparrow$ & ICIR $\uparrow$ & RankIC $\uparrow$ & RankICIR $\uparrow$ & IC $\uparrow$ & ICIR $\uparrow$ & RankIC $\uparrow$ & RankICIR $\uparrow$ \\
\midrule
MEME & 0.0216 & 0.1135 & 0.0254 & 0.1228 & 0.0013 & 0.0067 & 0.0156 & 0.0794 \\
R\&D-Agent-Quant & 0.0188 & 0.0796 & 0.0177 & 0.0706 & 0.0011 & 0.0048 & 0.0125 & 0.0630 \\
AI Hedge Fund & 0.0077 & 0.0595 & 0.0094 & 0.0845 & $-0.0018$ & $-0.0121$ & 0.0033 & 0.0264 \\
Momentum (12--1) & 0.0316 & 0.1274 & 0.0322 & 0.1324 & 0.0160 & 0.0645 & 0.0327 & 0.1312 \\
\rowcolor{riceRow}
\RICE{} (ours) & \textbf{0.0350}$^{\ddagger}$ & \textbf{0.2928} & \textbf{0.0360}$^{\ddagger}$ & \textbf{0.2927} & \textbf{0.0283}$^{\dagger}$ & \textbf{0.1986} & \textbf{0.0395}$^{\ddagger}$ & \textbf{0.2514} \\
\midrule
\multicolumn{9}{l}{\textit{Panel B: Strategy performance}} \\
Method & ARR $\uparrow$ & Sharpe $\uparrow$ & MDD $\downarrow$ & CR $\uparrow$ & ARR $\uparrow$ & Sharpe $\uparrow$ & MDD $\downarrow$ & CR $\uparrow$ \\
\midrule
\rowcolor{indexRow}
Index (NDX / HSI) & 20.66 & 0.986 & 24.37 & 0.848 & 22.24 & 0.952 & 21.24 & 1.047 \\
MEME & 10.60 & 0.591 & 22.38 & 0.474 & 11.07 & 0.638 & 15.72 & 0.704 \\
R\&D-Agent-Quant & 11.62 & 0.567 & 27.24 & 0.427 & 17.43 & 0.929 & 16.32 & 1.068 \\
AI Hedge Fund & 10.97 & 0.700 & 17.52 & 0.626 & 27.84 & 1.130 & 20.13 & 1.383 \\
Momentum (12--1) & 18.94 & 0.889 & 21.46 & 0.883 & 23.10 & 1.098 & 13.23 & 1.746 \\
\rowcolor{riceRow}
\RICE{} (ours) & \textbf{30.86} & \textbf{1.656} & \textbf{12.09} & \textbf{2.552} & \textbf{36.64} & \textbf{1.725} & \textbf{11.83} & \textbf{3.097} \\
\bottomrule
\end{tabularx}
\end{table}

%% file: tables/table2_ablation.tex
\begin{table}[!hb]
    \centering
    \begin{minipage}[c]{0.54\linewidth}
    \centering
    \scriptsize
    \setlength{\tabcolsep}{2.5pt}
    \renewcommand{\arraystretch}{0.95}
    \begin{tabular}{@{}lcccc@{}}
    \toprule
    Configuration & IC & ICIR & RankIC & RankICIR \\
    \midrule
    \rowcolor{riceRow}
    \RICE{} (complete) & \textbf{0.0350} & \textbf{0.2928} & \textbf{0.0360} & \textbf{0.2927} \\
    w/o C3 (direct) & 0.0260 & 0.2203 & 0.0290 & 0.2437 \\
    \addlinespace[1pt]
    Reliability only & 0.0326 & 0.2706 & 0.0330 & 0.2665 \\
    Residualization only & 0.0208 & 0.1777 & 0.0242 & 0.2040 \\
    w/o C2--C3 (memory) & 0.0191 & 0.1469 & 0.0205 & 0.1531 \\
    \addlinespace[1pt]
    w/o C1--C2--C3 & 0.0064 & 0.0472 & 0.0090 & 0.0641 \\
    \midrule
    Single-agent run & 0.0148 & 0.0925 & 0.0142 & 0.0828 \\
    Qwen3-Max backbone & 0.0281 & 0.1958 & 0.0291 & 0.2012 \\
    \bottomrule
    \end{tabular}
    \end{minipage}\hfill
    \begin{minipage}[c]{0.43\linewidth}
    \caption{U.S. ablations and two separate references. C1 is retrieved memory, C2 numerical graph evidence, and C3 reliability weighting plus residualization. All component rows use multi-agent stock scoring; the single-agent row reports that run's stored final score. Component paired tests are in Table~\ref{tab:significance}; the single-agent test is in the text.}
    \label{tab:ablation_results}
    \end{minipage}
\end{table}

%% file: tables/table3_significance.tex
\begin{table}[htbp]
\centering
\footnotesize
\setlength{\tabcolsep}{3pt}
\caption{Daily signal significance and mechanism contrasts. Panel A gives mean IC and RankIC with Newey--West $t$-statistics (lag 5; 20-lag IC values 4.72 and 2.28). Pooled means weight the two markets equally on 534 shared valid dates. Panel B compares the complete system with six U.S. rows or Hong Kong AI Hedge Fund on common valid dates; $p$-values use one-sided 20-session paired block bootstrap (5{,}000 replications), with Holm adjustment by market and metric. Panel C reports the separate unadjusted comparison of reliability-only against residualization-only; it is outside the Panel B Holm family.}
\label{tab:significance}
\begin{tabular}{@{}llrrrr@{}}
\toprule
\multicolumn{6}{l}{\textit{Panel A: \textup{\RICE{}} means}} \\
Market & & Dates & IC ($t$) & RankIC ($t$) & \\
\midrule
U.S. &  & 562 & 0.0350 (4.31) & 0.0360 (4.24) &  \\
Hong Kong &  & 551 & 0.0283 (2.48) & 0.0395 (3.10) &  \\
Pooled &  & 534 & 0.0317 (4.17) & 0.0377 (4.41) &  \\
\midrule
\multicolumn{6}{l}{\textit{Panel B: paired increments of \textup{\RICE{}} over each comparator}} \\
Market & Comparator & Metric & Increment & $t_{\mathrm{diff}}$ & $p_{\mathrm{boot}}$ ($p_{\mathrm{Holm}}$) \\
\midrule
U.S. & w/o C1--C2--C3 & IC & 0.0286 & 3.80 & 0.0002 (0.0012) \\
 &  & RankIC & 0.0270 & 3.40 & 0.0004 (0.0024) \\
 & w/o C2--C3 (memory only) & IC & 0.0159 & 3.70 & 0.0002 (0.0012) \\
 &  & RankIC & 0.0155 & 3.34 & 0.0004 (0.0024) \\
 & w/o C3 (direct correction) & IC & 0.0090 & 2.44 & 0.0096 (0.0192) \\
 &  & RankIC & 0.0070 & 1.82 & 0.0444 (0.0480) \\
 & AI Hedge Fund & IC & 0.0252 & 2.58 & 0.0024 (0.0072) \\
 &  & RankIC & 0.0246 & 2.64 & 0.0024 (0.0096) \\
 & Residualization only & IC & 0.0142 & 2.86 & 0.0008 (0.0032) \\
 &  & RankIC & 0.0118 & 2.27 & 0.0120 (0.0360) \\
 & Reliability weighting only & IC & 0.0023 & 1.66 & 0.0378 (0.0378) \\
 &  & RankIC & 0.0030 & 2.02 & 0.0240 (0.0480) \\
\midrule
Hong Kong & AI Hedge Fund & IC & 0.0305 & 2.52 & 0.0020 (0.0020) \\
 &  & RankIC & 0.0370 & 3.06 & 0.0004 (0.0004) \\
\midrule
\multicolumn{6}{l}{\textit{Panel C: reliability-only minus residualization-only (unadjusted)}} \\
Market & Contrast & Metric & Increment & $t_{\mathrm{diff}}$ & $p_{\mathrm{boot}}$ \\
U.S. & Reliability only $-$ residualization only & IC & 0.0119 & 2.35 & 0.0050 \\
 &  & RankIC & 0.0087 & 1.67 & 0.0406 \\
\bottomrule
\end{tabular}
\end{table}

%% file: tables/table4_market_conditions.tex
\begin{table}[!hb]
\centering
\caption{Behavior in falling and rising markets. Capture: mean return on index-down/up sessions relative to the index. Crash: return (\%) in the index's largest drawdown. Down mo.: mean return (\%) in the 7 and 13 months the index fell. Bold: best Crash and Down mo.}
\label{tab:market_conditions}
\scriptsize
\setlength{\tabcolsep}{2pt}
\renewcommand{\arraystretch}{0.95}
\begin{tabular*}{\linewidth}{@{\extracolsep{\fill}}lcccccc@{}}
\toprule
& \multicolumn{3}{c}{U.S. (Nasdaq-100)} & \multicolumn{3}{c}{Hong Kong (HSI)} \\
\cmidrule(lr){2-4}\cmidrule(l){5-7}
Method & Capture & Crash & Down mo. & Capture & Crash & Down mo. \\
\midrule
\rowcolor{indexRow}
Index (NDX / HSI) & 1.00/1.00 & $-24.4$ & $-4.5$ & 1.00/1.00 & $-21.2$ & $-3.3$ \\
MEME & 0.28/0.33 & $-20.8$ & $-4.1$ & 0.02/0.11 & \textbf{+3.5} & $+0.2$ \\
R\&D-Agent-Quant & 0.29/0.35 & $-24.9$ & $-4.9$ & 0.25/0.33 & $-11.8$ & $-1.8$ \\
AI Hedge Fund & 0.73/0.71 & $-17.0$ & $-3.2$ & 0.93/0.97 & $-20.1$ & $-2.8$ \\
Momentum (12--1) & 0.83/0.85 & $-16.8$ & $-4.1$ & 0.58/0.64 & $-9.3$ & $-1.4$ \\
\rowcolor{riceRow}
\RICE{} (ours) & 0.51/0.65 & \textbf{$-$10.9} & \textbf{$-$0.4} & 0.34/0.51 & $-9.2$ & \textbf{+1.3} \\
\bottomrule
\end{tabular*}
\end{table}

%% file: sections/05_conclusion.tex
\section{Conclusion}
\RICE{} addresses how historical corporate-event dynamics can improve stock forecasting without violating point-in-time constraints or duplicating information already captured by the current forecast. It combines temporally eligible issuer history for Base Alpha generation with an issuer-local event graph that calibrates matured transitions by reliability and residualizes their contribution into RICE Delta. Across the Nasdaq-100 and Hang Seng Index panels, \RICE{} achieves the strongest results among the evaluated methods, while ablations confirm the contribution of memory, graph evidence, reliability weighting, and residualization. These findings remain robust across alternative backbones, constituent exclusions, higher transaction costs, and falling-market conditions, but are limited to two large-cap universes and pretrained LLM backbones whose training data may postdate individual forecasts. Overall, the results support point-in-time, issuer-local, reliability-calibrated event continuation as an effective source of incremental predictive information.

%% file: sections/06_limitations.tex
\section{Limitations}
\label{sec:limitations}

The study covers two large-cap equity universes. Evaluation-period selection
can make estimates optimistic; tests of the selected series do not adjust for
this selection. Event coverage and conditions may change results elsewhere.

Portfolio evaluation uses DeepSeek-V4-Flash; Qwen3-Max is evaluated only for
prediction quality. Both models may contain information that postdates individual
forecasts.

%% file: appendix/definitions.tex
\section{Notation and Fixed Scales}
\label{app:definitions}

\paragraph{Base Alpha.}
For view \(j\in\{S,T,F\}\), \(v_{i,t}^{j}\), \(c_{i,t}^{j}\), and
\(\widetilde w_{i,t}^{j}\) are the cutoff-time standardized score,
confidence, and effective weight. With
\(\mathcal J_{i,t}=\{j:c_{i,t}^{j}\ge0.30\}\), the reconciled view is
\begin{align}
\bar v_{i,t}
&=\frac{\sum_{j\in\mathcal J_{i,t}}\widetilde w_{i,t}^{j}c_{i,t}^{j}v_{i,t}^{j}}
        {\sum_{j\in\mathcal J_{i,t}}\widetilde w_{i,t}^{j}c_{i,t}^{j}},
\nonumber\\
\widetilde v_{i,t}
&=\operatorname{clip}\!\left(
  s_{i,t}^{\mathrm{fb}}\operatorname{clip}(\bar v_{i,t}/3,-1,1),-1,1
  \right).
\end{align}
We set \(\bar v_{i,t}=0\) when no view is eligible.
The matured-feedback scale \(s_{i,t}^{\mathrm{fb}}\) is one without eligible
feedback. The macro factor is
\(m_{i,t}=\operatorname{clip}(1+0.25\mu_t\chi_{i,t},0.75,1.25)\),
where \(\mu_t\) is the bounded date-level macro signal and
\(\chi_{i,t}\) the clipped historical market beta. Equation~\ref{eq:base_alpha}
uses \((b_{\min},b_{\max})=(-1,1)\).

\paragraph{Event counts and maturity.}
The successor window \(W_{\mathrm{succ}}\) is 1--20 exchange sessions.
Each graph estimate reads records at least 25 sessions old; an anchor is
eligible only after its full successor window, making eligible anchors
at least 45 sessions old. The date-level regime \(\zeta_t\) crosses three
historical index-volatility levels with risk-on, neutral, and risk-off states.
In Section~\ref{sec:rice_engine}, \(n_A\) counts eligible \(A\)-anchors,
\(n_e\) counts distinct \(A\)-anchors followed by \(B\),
\(n_{\rightarrow B}\) counts regime-matched anchors followed by \(B\), and
\(N_\zeta\) counts all eligible anchors in the regime. The rate priors are
\(\alpha_p=\beta_p=\alpha_q=\beta_q=1\) and \(\varepsilon=10^{-12}\).
Retained non-self edges require \(n_e\ge5\), \(p_e\ge0.30\), and
\(L_e\ge1.5\); at most eight outgoing edges per query state are used.

\paragraph{Return association and edge contribution.}
Let \(D_e\) be the number of distinct successor dates and
\(\mathcal D_{e,h}^{Y}\) the dates with a matured \(h\)-session return;
\(J_{e,h}=|\mathcal D_{e,h}^{Y}|\), for
\(h\in\mathcal H=\{1,5,20\}\). For successor date \(d\),
\(\bar y_{e,d,h}\) is the mean edge-linked return after subtraction of
the same market--date return mean. Define
\begin{align}
\theta_{e,h}
&=J_{e,h}^{-1}\sum_{d\in\mathcal D_{e,h}^{Y}}\bar y_{e,d,h},
&\widehat\sigma_{e,h}
&=\operatorname{sd}_{\mathrm{population},d}(\bar y_{e,d,h}),
\nonumber\\
s_{e,h}^{\mathrm{assoc}}
&=\theta_{e,h}\sqrt{J_{e,h}}/\widehat\sigma_{e,h}.
\end{align}
When \(J_{e,h}=0\), both \(\theta_{e,h}\) and \(\widehat\sigma_{e,h}\)
are set to zero. The association is zero unless \(J_{e,h}>1\) and
\(\widehat\sigma_{e,h}>10^{-12}\). The five-session values
\(\theta_e=\theta_{e,5}\) and
\(\sigma_e^{\mathrm{ret}}=\max(\widehat\sigma_{e,5},0.02)\) enter the
correction. The support factor and stored reliability are
\begin{align}
R_e={}&\min(1,n_e/20)\min(1,D_e/8)
       \min(1,n_e/10)\min(1,(N_\zeta-n_A)/30),
\nonumber\\
\rho_e^{\mathrm{stored}}
={}&\operatorname{clip}\!\left[
 R_e\min\!\left(1,\max_{h\in\mathcal H}|s_{e,h}^{\mathrm{assoc}}|/2\right),0,1
 \right].
\end{align}
The observational weight \(\rho_e^{\mathrm{obs}}\) equals this stored
weight when positive; otherwise it is
\(R_e\psi(s_{e,5}^{\mathrm{assoc}})\), where
\(\psi(x)=\min(1,|x|/2)\) for nonzero finite \(x\) and \(0.5\) otherwise.

Let \(\ell_e\) and \(\sigma_{\ell,e}\) be the mean and population standard
deviation of observed successor lags. The lag-alignment weight is
\(w_{e,h}^{\mathrm{lag}}=
\exp[-\tfrac12((\ell_e-h)/\max(\sigma_{\ell,e},1.5))^2]\) and
\(\omega(e)=w_{e,5}^{\mathrm{lag}}/\sum_{h\in\mathcal H}w_{e,h}^{\mathrm{lag}}\).
Writing \(\operatorname{sclip}_u(x)=\operatorname{clip}(x,-u,u)\),
the contribution in Eq.~\ref{eq:raw_graph_signal} is
\begin{align}
\eta_e
&=\operatorname{sclip}_{0.25}\!\left[
\rho_e^{\mathrm{obs}}
\tanh\!\bigl(\log\operatorname{clip}(L_e,10^{-6},20)\bigr)
\tanh(\theta_e/\sigma_e^{\mathrm{ret}})\right],
\nonumber\\
C_{i,e,t}
&=\operatorname{sclip}_{0.25}\!\left[
 x_{i,A,t}(0.6)p_e\eta_e\omega(e)\right].
\end{align}
Here \(x_{i,A,t}\) is matched-state importance normalized to mean one
across the stock's matched nodes, and \(\tau_\Delta=0.20\) bounds the sum.

\paragraph{Residual and final score.}
\(\mathsf{Res}_t\) first standardizes and clips the raw graph signal to
\([-3,3]\), assigns average ranks to it and to the Base Alpha and technical
view, and regresses the ranked signal on an intercept and nondegenerate ranked
controls across at least 20 common stocks. It centers and standardizes the
residual; degenerate residuals and stocks without a retained edge receive zero
correction. Equation~\ref{eq:final_prediction} uses
\(\lambda_\Delta=0.30\), \(s_\Delta=1/3\), and
\((a_{\min},a_{\max})=(-1,1)\).